# The Resale Price Prediction of Secondhand Jewelry Items Using a Multi-modal Deep Model with Iterative Co-Attention

Yusuke Yamaura
Research & Technology Group
Fuji Xerox Co., Ltd.
yamaura.yusuke@fujixerox.co.jp

Nobuya Kanemaki
Planning Department AI Business Development Unit
Nishikawa Communications Co. Ltd.
kanemaki@nishikawa.jp

Yukihiro Tsuboshita
Research & Technology Group
Fuji Xerox Co., Ltd.
Yukihiro.Tsuboshita@fujixerox.co.jp

## ABSTRACT

The resale price assessment of secondhand jewelry items relies heavily on the individual knowledge and skill of domain experts. In this paper, we propose a methodology for reconstructing an AI system that autonomously assesses the resale prices of secondhand jewelry items without the need for professional knowledge. As shown in recent studies on fashion items, multimodal approaches combining specifications and visual information of items have succeeded in obtaining fine-grained representations of fashion items, although they generally apply simple vector operations through a multimodal fusion. We similarly build a multimodal model using images and attributes of the product and further employ state-of-the-art multimodal deep neural networks applied in computer vision to achieve a practical performance level. In addition, we model the pricing procedure of an expert using iterative co-attention networks in which the appearance and attributes of the product are carefully and iteratively observed. Herein, we demonstrate the effectiveness of our model using a large dataset of secondhand no brand jewelry items received from a collaborating fashion retailer, and show that the iterative co-attention process operates effectively in the context of resale price prediction. Our model architecture is widely applicable to other fashion items where appearance and specifications are important aspects.

## KEYWORDS

Multimodal, Deep Learning, Co-Attention, Secondhand Fashion Item, Price Prediction text here

## 1 INTRODUCTION

The fashion resale market has been rapidly growing and is becoming a larger part of the fashion industry owing to the prevalence of BtoC e-commerce sites and CtoC flea market apps used by consumers. A recent increase in secondhand shoppers is also driving this growth [1, 2]. Owing to such increase in demand and acceleration in trading, an appropriate price assessment of secondhand fashion items has become a major issue for retailers. Particularly in the case of jewelry items, a more proper price assessment is required because their unit price is much higher than that of other fashion items. Besides, the resale pricing of secondhand jewelry items heavily relies on the individual knowledge and skill of domain experts with several years of practical experience in purchasing, and jewelry retailers are therefore imposed with large costs for educating workers. To resolve this problem, through a partnership with Japan's largest fashion resale retailer, we are constructing an AI system that autonomously assesses the resale prices of secondhand jewelry items without professional knowledge, even by laymen.

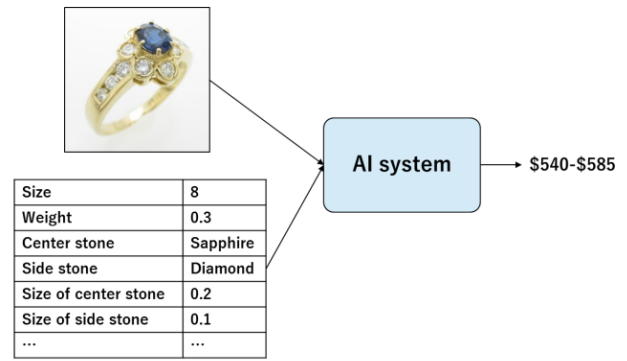

**Figure 1: The summary of the resale price prediction task of secondhand jewelry items. AI system takes an image and attributes of a product as input and suggest users a relevant resale price.**

Most predictive tasks basically rely on structured data [3, 4], such as the attributes or specifications. However, regarding the fashion items, their design appearance is extremely important in terms of their characterization, and it is difficult to estimate their prices correctly using their specifications alone. In recent studies on fashion items, product images are often utilized to incorporate visual information into their analytics. For a fashion product search or retrieval, an intuitive search [5-8] is realized through multimodal representations combining product images and their textual descriptions or attributes. In terms of fashion product recommendations, visual features of the product images are integrated with the item descriptions as latent factors to improve the recommendation performance [9-11]. These previous studies have shown that visuality is an essential factor in the analytics of fashion products, and multimodal approaches are effective at obtaining a better representation of fashion products. However, previous multimodal models generally employ traditional vector operations in integrating the feature vectors of different modalities, such as a simple concatenation, element-wise addition, or element-wise product. They are thus insufficient for acquiring representations of high-level associations between modalities.

At the same time, various multimodal fusion methods have been well studied regarding Visual Question Answering (VQA) tasks [12], which are typical multimodal challenges in the field of computer vision. Given an image and a natural language question regarding the image, the task is to derive an answer to the question. Bilinear models have shown an outstanding performance in VQA [13-18] because they enable the learning of high-dimensional interactions between modalities by encoding full second-order products. In fact, they are approximated or simplified to reduce the number of parameters owing to the main issue of a huge dimensionality; however, they still provide better multimodal representations and predictive performances than classical vector combinational approaches. As another way to improve their accuracy, co-attention approaches have recently been proposed [15, 19-22]. The co-attention mechanism enhances important regions of the input image and words of the question and enables to boost the predictive performance. Attention maps are calculated from interactions of the modalities, and are helpful in understanding the reasons for the prediction results; thus, they are often utilized in explanations and reasoning tasks [20, 22].

Similar to the previous studies on fashion items, we also employ a multimodal model to combine a product image and its attributes, while further employing state-of-the-art methods in multimodal deep learning to achieve a practical level of accuracy regarding the resale pricing of secondhand fashion items. To demonstrate the performance of our model, we conducted an extensive experiment using a large dataset of purchasing and selling information of secondhand no brand rings received from a collaborative fashion retailer. This dataset is challenging because the pricing process of no brand rings is slightly different from brand items, whose product type has a strong impact on their price assessment. For no brand rings, the design appearance and specifications are more important because they have no product type to be compared to. In a practical price assessment of no brand rings by an expert, the appearance and attributes of the product are carefully and iteratively observed. We attempted to reproduce the pricing procedure of an expert using multimodal deep neural networks to achieve a practical price assessment. We employ a co-attention mechanism to enable focusing on important parts of the image and attributes and additionally propose a new iterative architecture for repeating the co-attention process.

The primary contributions of our research are as follows.

- Through collaboration with Japan's largest resale retailer, we worked on achieving a highly practical task, namely, the resale price assessment of secondhand jewelry items.
- Herein, we present an iterative co-attention network using state-of-the-art methods in multimodal deep learning, which achieves a practical performance level.

The rest of the paper is organized as follows. Section 2 reviews the related works. Section 3 formally explains the resale price prediction task and Section 4 describes our proposed network architectures. Section 5 experimentally evaluates our model with a large dataset of secondhand no brand rings, where we discuss the effectiveness of our approaches. We finally conclude the paper in Section 6.

## 2 RELATED WORK

Multimodal approaches have recently become popular for use in several fashion domain tasks because they provide more fine-grained representations of fashion products. Regarding fashion product searches and retrieval, Zhao et al. [5] introduced a multimodal search method with a clothing image and its attributes as a query and enabled to transform appearance of the clothing to desirable one by manipulating the attributes. Liao et al. [6] utilized multi-level semantics extracted from a query image and user's feedback for them for more precise search. Tautkute et al. [7] proposed a multimodal style search engine based on the similarity of aesthetics and style, using product images and text descriptions. They deployed it to their publicly available web application where users can search furniture products by selecting interior design patterns. Shanker et al. [8] used deep and shallow convolutional neural network to extract both high and low level features from a clothing image. They deployed their model to one of largest fashion e-commerce site in India. In terms of fashion product recommendation, He et al. [9, 10] utilized visual features extracted from a product image into Matrix Factorization to improve their recommendation performance and also alleviate cold start issues. Liu et al. [11] incorporated style and categorical features extracted from a product image. Previous studies on fashion product searches, retrieval, and recommendations have shown that incorporating product images provides information of the design appearance, user preference, and semantics, which do not appear in other modalities. Furthermore, multimodal approaches have proven to be an effective method for a fashion item analysis.

By contrast, multimodal fusion strategies have been well studied regarding VQA tasks [12], which are typical multimodal challenges in the field of computer vision. Bilinear models have shown to outperform traditional linear models in VQA because they consider high-dimensional interactions between modalities by computing the outer-product of the feature vectors. The main issue is the significant amount of computational resources required owing to an explosion of combinations of the feature vectors when they have high dimensionality. Therefore, current approaches approximate or simplify the bilinear models to reduce the number of parameters. Fukui et al. [13] applied a count sketch method to reduce the parameters. Kim et al. [14] approximates the bilinear interactions by Hadamard product with low-rank constraint. Yu et al. [15] proposed MFB model and imposed Matrix Factorization on sliced matrices of a bilinear tensor. They extended their approach to a higher-order setting by MFH, consists of cascaded MFB blocks [16]. Younes et al. [17] introduced Tucker Decomposition framework to the bilinear tensor with rank sparsity constraint and their recent work, BLOCK [18] similarly applied Block Term Decomposition. Bilinear-based models have shown an outstanding performance in terms of VQA; nevertheless, they have not been frequently employed in practical applications.

As another way to improve multimodal models, co-attention mechanisms have been introduced in VQA [15, 19- 22]. Such co-attention enhances the important regions of the input image and words of the question representations and helps improve the predictive performance. Furthermore, the stacking of co-attention blocks has shown to refine the representations and increase their accuracy. Lu et al. [19] proposed a hierarchical co-attention model considering different level of sentence and presented basic logics of parallel and alternative co-attention mechanisms. Nam et al. [20] exploited multi-step dual attention networks for reasoning and matching. Nguyen and Okatani [21] proposed dense co-attention networks which computes attentions from bilinear interactions between modalities. Younes et al. [22] presented MuRel cell which models pair-wise relations between each region of the image and question and sequentially connected them. These iterative co-attention models allow more fine-grained representations to be obtained, and boost the predictive performance. However, from the viewpoint of neuroscience, they do not follow the human attention system [23, 24]; in addition, there have been few studies on applying a co-attention framework in practical applications.

We employed such state-of-the-art methods in VQA tasks to construct a multimodal model for the resale price assessment of secondhand jewelry items. Moreover, we modeled an expert's pricing procedure through which the appearance and specifications of a product are carefully and iteratively observed using iterative co-attention networks obedient to the human attention system, and clarified important factors of the multimodal inputs by visualizing attention maps to help users understand the reasons for the predictive results.

## 3 RESALE PRICE PREDICTION

A summary of a resale price prediction task of secondhand jewelry items is illustrated in Figure 1. Given an image and attributes of a product as the input, the multimodal model predicts its resale price. We construct the model using deep neural networks in a supervised learning manner, providing the selling prices as the labels. Strictly speaking, actual market prices may not be uniquely determined because there are various external factors influencing them, such as the economic conditions, market trends of the metals, and encounters with customers. In the present study, we regard the selling prices as the market prices reflecting the various factors described above, and use the selling prices as the teaching values for the prediction.

We could formulate our price prediction problem as a regression task but then the predicted price value would be some unique value although the market prices are not uniquely determined due to the various external factors as we mentioned above. We therefore formulate our problem as a classification task since it naturally provides multiple candidate prediction values, where each category has a confidence value.

## 4 MODEL ARCHITECTURE

In this section, we provide the details of our model architecture. We employ state-of-the-art methods in multimodal deep learning and model the expert's pricing procedure using iterative co-attention mechanisms in which the appearance and specifications of the product are carefully and iteratively observed. The entire network, as illustrated in Figure 2, consists of sequential co-attention cells, each being responsible for a refinement of the representations of the image and attributes using a co-attention mechanism and fusion to provide a predictive feature vector. Using an iterative co-attention process, the predictive feature vectors are accumulated to form the final price decision, and the resale price is gradually determined through careful and iterative observations of the appearance and specifications. We first explain the initial representations of the image and attributes in Section 4.1 and 4.2 respectively, and then describe the details of co-attention cell in Section 4.3. We additionally introduce an ensemble strategy in Section 4.4.

### 4.1 IMAGE REPRESENTATION

As in many previous studies on multimodal deep learning using an image as one of modalities. We employ ResNet-50 model [25] to extract visual feature maps from an input image. If more fine-grained visual representations are required, it can be replaced by deeper ones such as ResNet-101 and 152 or other state-of-the-art models, but we use ResNet-50 due to the limitation of GPU memory. The input images are size of 448×448 and we use feature maps of 2048×14×14 dimensions from 'res5c' block before the final average pooling layer in ResNet-50. We denote dimensions of feature maps as $d_v \times D$ so $d_v = 2048$ and $D = 14 \times 14$. The network is pre-trained with ImageNet [26] dataset and fine-tuned with jewelry images in an end-to-end manner.

### 4.2 ATTRIBUTE REPRESENTATION

To encode the attributes, we apply two-layered embedding networks, as illustrated in Figure 2. The input attributes include both continuous and categorical variables. The categorical variables are assigned IDs and are transformed into one-hot vectors beforehand. The first embedding layer has multiple fully connected layers, each for a particular categorical attribute, and converts the one-hot vectors into distributed representations such as word embedding [27]. The number of units $u$ for each fully connected layer is determined using $u = ceil(\tilde{d}/2)$, where $\tilde{d}$ indicates the dimensions of a one-hot vector. The second embedding layer also consists of fully connected layers, each being for a particular categorical and continuous attribute, and projects the distributed representations into a high-dimensional space $d_a$ common to all attributes, which is set to 200. The output feature vectors are stacked in a feature map of $d_a \times T$ dimensions, where $T$ is the number of attribute types.

### 4.3 CO-ATTENTION CELL

We now describe the details of co-attention cell; see Figure 3. It takes visual and attributes feature maps, $h_v \in \mathbb{R}^{d_v \times D}$ and $h_a \in \mathbb{R}^{d_a \times T}$ respectively and outputs their refined representations as well as a predictive feature vector $h_o \in \mathbb{R}^{d_o}$. We denote these visual and attribute representations into the $(l+1)$-th co-attention cell as $h_v^{(l)}$ and $h_a^{(l)}$, respectively and output predictive feature vector as $h_o^{(l)}$ where $l \in [0, L-1]$ and $L$ is the number of cells. For the initial representations before the first co-attention cell are denoted as $h_v^{(0)}$ and $h_a^{(0)}$, where $l = 0$. All co-attention cells share their weight parameters inside them.

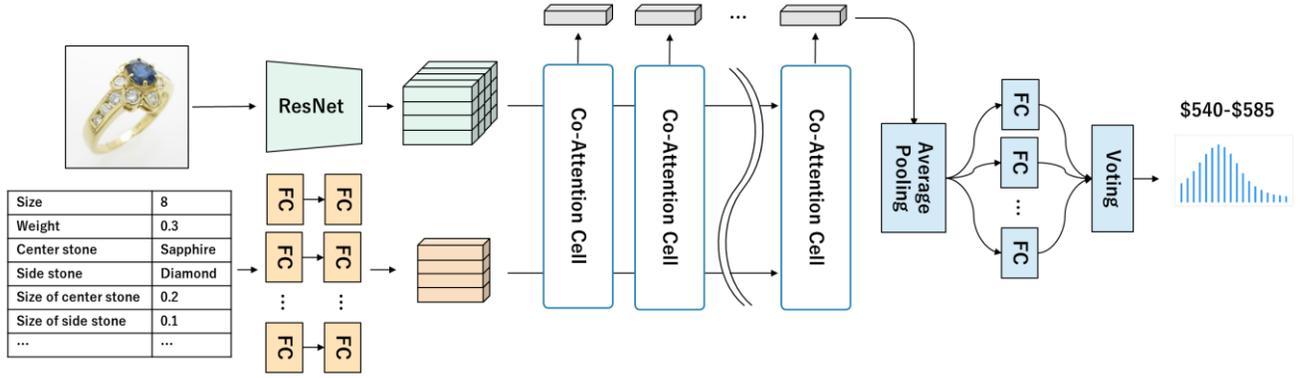

Figure 2: An overview of our iterative co-attention networks. An image and the attributes of a product are fed to the network and each feature map is initially extracted. A single co-attention cell takes these representations to enhance their important parts using a co-attention mechanism and outputs a predictive feature vector. Using an iterative co-attention process through the sequential cells, the visual and attribute representations are repeatedly refined, and predictive feature vectors are accumulated to form the final price decision. In the final prediction step, we employ a coarse-to-fine ensemble strategy by voting by multiple classifiers. FC in the figure means the fully connected layer.

*4.3.1 Top-down and bottom-up attentions.* We model the expert's pricing procedure using iterative co-attention networks in which the appearance and specifications of the product are carefully and iteratively observed. From the viewpoint of neuroscience, an iterative co-attention process has the potential for modeling the human attention system. Top-down and bottom-up attention are two human visual attention systems [23, 24]; top-down attention is intentionally caused by specific stimulus in the context of the current task, and bottom-up attention is passively drawn to an unexpected stimulus. When a human acts in the real world, the top-down and bottom-up attention systems do not work individually, but both are applied interactively. Top-down attention can be expressed as 'attribute-to-visual' attention because the appearance is observed based on the specific purpose with the provided attributes. By contrast, bottom-up attention can be considered 'visual-to-attribute' attention because visual saliency provides further interest regarding the attributes. Based on these findings, we propose formulating the human attention system in an iterative co-attention network as follows.

Given the visual and attribute representations, $h_v^{(l)}$ and $h_a^{(l)}$, they are firstly summed to calculate attentions as

$$\hat{h}_v^{(l)} = \sum_{i=1}^{D} h_v^{(l)}, \quad (1)$$

$$\hat{h}_a^{(l)} = \sum_{i=1}^{T} h_a^{(l)}. \quad (2)$$

These $\hat{h}_v^{(l)} \in \mathbb{R}^{d_v}$ and $\hat{h}_a^{(l)} \in \mathbb{R}^{d_a}$ are the feature vectors for each modality. Top-down and bottom-up attention maps, $A_v^{(l)} \in \mathbb{R}^{d_v \times D}$ and $A_a^{(l)} \in \mathbb{R}^{d_a \times T}$, are each calculated by

$$A_v^{(l)} = \text{Softmax}\left(f\left(h_v^{(l)}, \hat{h}_a^{(l)}\right)\right) \quad (3)$$

and

$$A_a^{(l)} = \text{Softmax}\left(f\left(\hat{h}_v^{(l+1)}, h_a^{(l)}\right)\right) \quad (4)$$

respectively. Note that $f(\cdot)$ is a common multimodal fusion operation between both attentions, which learn high-dimensional associations between the modalities and specify the factors to be enhanced. We discuss $f(\cdot)$ in next subsection in detail. We calculate the bottom-up attention using refined visual representations. The refined representations are calculated by element-wise multiplication defined as

$$h_v^{(l+1)} = A_v^{(l)} * h_v^{(l)}, \quad (5)$$
$$h_a^{(l+1)} = A_a^{(l)} * h_a^{(l)}. \quad (6)$$

These representations become next input to followed co-attention cell. We finally perform multimodal fusion inside the cell to derive a predictive feature vector from attended representations by

$$h_o^{(l)} = g\left(\hat{h}_v^{(l+1)}, \hat{h}_a^{(l+1)}\right). \quad (7)$$

where $g(\cdot)$ is also a multimodal fusion operation but distinguished from $f(\cdot)$ because the expressive power required is quite different between calculation of attention maps and calculation of predictive feature vector and also there is a limitation of GPU memory.

By iterative co-attention process, we obtain multiple predictive feature vectors $h_o^{(l)}$ for $l = 0, ..., L - 1$. This accumulated feature vectors could be considered to form the final price decision so we compute the average of them as follows:

$$\hat{h}_o = \frac{1}{L}\sum_{l=0}^{L-1} h_o^{(l)}. \quad (8)$$

The averaged predictive feature vector is fed to an ensemble module explained in Section 4.4 in detail.

Previous sequential co-attention methods [19-22] have not followed the theory of neuroscience and have applied the two attention process at the same time. In contrast, our model applies the two attention processes alternately, following theory.

*4.3.2 Fusion strategy.* The co-attention cell requires multimodal fusion for three times, two for calculation of attention maps and the rest is for calculation of predictive feature vector, denoted as $f(\cdot)$ and $g(\cdot)$ respectively. We employ the MFB model [15] for $f(\cdot)$. It has appealing expressiveness in spite of drastic reduction in number of parameters. We briefly review the logics of MFB model. For simplicity, feature vectors for different two modalities are denoted as $x \in \mathbb{R}^{d_x}$ and $y \in \mathbb{R}^{d_y}$ below. Bilinear model of two representations is defined as follows:

$$z_i = x^T W_i y, \tag{9}$$

where $W_i \in \mathbb{R}^{d_x \times d_y}$ is a projection matrix and $z_i \in \mathbb{R}$ is the output of bilinear model. The matrix $W_i$ can be factorized into two low-rank matrices by Matrix Factorization method as

$$z_i = x^T U_i V_i^T y = \sum_{d=1}^{k} x^T u_d v_d^T y = \mathbb{I}^T (U_i^T x \circ V_i^T y), \tag{10}$$

where $U_i \in \mathbb{R}^{d_x \times k}$ and $V_i \in \mathbb{R}^{d_y \times k}$ are the latent matrices, $\circ$ is the Hadamard product of two vectors, $\mathbb{I} \in \mathbb{R}^k$ is an all-one vector. The matrix $W_i$ is a sliced matrix of the bilinear tensor and we need to obtain $z \in \mathbb{R}^o$. Therefore, two latent matrices are extended to three-order tensors, $U \in \mathbb{R}^{d_x \times k \times o}$ and $V \in \mathbb{R}^{d_y \times k \times o}$ respectively. We further reformulate these tensors to 2-D matrices, $\widetilde{U} \in \mathbb{R}^{d_x \times ko}$ and $\widetilde{V} \in \mathbb{R}^{d_y \times ko}$. Accordingly, $z$ can be rewritten as follows:

$$z = \text{SumPooling}(\widetilde{U}^T x \circ \widetilde{V}^T y, k), \tag{11}$$

where the function $\text{SumPooling}(s, k)$ means sum pooling over $s$ with window size $k$.

The MFB model mainly consists of expand stage and squeeze stage and consequently requires only embedding layer for each modality, represented as $\widetilde{U}$ and $\widetilde{V}$ above. To calculate our attention maps, we additionally perform feature transformation with 1-D convolutions and ReLU activation as depicted in Figure 3. We set depth dimension of first convolution output to 128 and second to 1, which means that we provide one glimpse in one attention process. Therefore, the function $f(\cdot)$ also includes these transformations as well as fusion. For the sake of simplicity we denote $f(\cdot)$ as

$$f = \text{MFB}(x, y). \tag{12}$$

On the other hand, we require high-level expressiveness for calculation of predictive feature vectors. According to our preliminary experiments to compare several fusion methods, we find BLOCK model [18] demonstrates outstanding performance and it enables us to control the expressivity and complexity of the model. Thus, we define $g(\cdot)$ as follows:

$$g = \text{BLOCK}(x, y). \tag{13}$$

We also briefly review the core part of BLOCK model.

The bilinear model takes two vectors $x \in \mathbb{R}^{d_x}$ and $y \in \mathbb{R}^{d_y}$, and projects them as follows:

$$z = \mathcal{T} \times_1 x \times_2 y, \tag{14}$$

where $z \in \mathbb{R}^o$ is an output feature vector and $\mathcal{T} \in \mathbb{R}^{d_x \times d_y \times o}$ is a bilinear tensor. The bilinear tensor is expressed in terms of Block Term Decomposition as

$$\mathcal{T} := \sum_{r=1}^{R} \mathcal{D}_r \times_1 A_r \times_2 B_r \times_3 C_r, \tag{15}$$

where $\forall r \in [1, R]$, $\mathcal{D}_r \in \mathbb{R}^{L \times M \times N}$ is the $r$-th core tensor, $A_r \in \mathbb{R}^{d_x \times L}$, $B_r \in \mathbb{R}^{d_y \times M}$, and $C_r \in \mathbb{R}^{o \times N}$ are the $r$-th latent matrices. It can be written as

$$\mathcal{T} = \mathcal{D}^{bd} \times_1 A \times_2 B \times_3 C, \tag{16}$$

where $A = [A_1, \ldots, A_R]$ (same for $B$ and $C$), and $\mathcal{D}^{bd} \in \mathbb{R}^{LR \times MR \times NR}$ is the block-superdiagonal tensor of $\{\mathcal{D}_r\}_{1 \leq r \leq R}$.

The BLOCK model consequently approximates a full bilinear tensor by a sum of low-rank terms composed of $A_r$, $B_r$, and $C_r$, and $\mathcal{D}_r$ (the tensor size is controllable), and imposes rank-sparsity constraints. However, we do not apply rank-sparsity because our computational resources have the capability of computing the core tensors $\mathcal{D}_r$. We employ the BLOCK model for the final fusion inside the co-attention cell, and therefore our single co-attention cell is composed of MFB-based top-down and bottom-up attention mechanisms and a BLOCK fusion module.

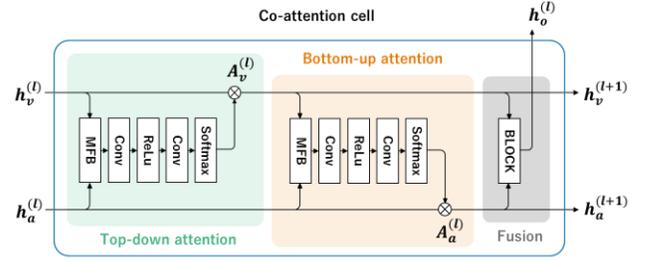

Figure 3: Details of the single co-attention cell. It takes visual and attribute representations, $h_v^{(l)}$ and $h_a^{(l)}$, as input and then outputs refined them, $h_v^{(l+1)}$ and $h_a^{(l+1)}$, and a predictive feature vector $h_o^{(l)}$. Inside the cell, top-down and bottom-up attention are calculated by MFB fusion method. The final fusion is performed by BLOCK fusion model.

## 4.4 MULTIPLE CLASSIFIER ENSEMBLE

We also introduce an ensemble strategy, Multiple Classifier Ensemble (MCE), into our model for further improvement. As described in Section 3, we formulate the resale price prediction problem as a classification task. Fine-grained classification becomes difficult when the amount of data belong to each class is small. To solve this problem, we set multiple classifiers with coarse labels shifting the price zones and apply a voting based on their multiple prediction results. This coarse-to-fine approach increases the amount of data belong to each class and results in more stable prediction through a voting strategy. The final prediction result is determined by voting over the original labels, and a softmax function is applied to obtain the class probability distribution. In order to apply our coarse-to-fine ensemble strategy, we additionaly define hyper-parameters, the number of classifiers $N_c$, the number of coarse classes $\#\{C_k\}$ ($C_k$ means class sets defined for $k$-th classifier), and shifting price band $b$ (in dollar). The number of coarse classes $\#\{C_k\}$ should be set less than the number of classes in final classification (18 classes in our case as explained in Section 5.1). We empirically determined $N_c = 20$, $\#\{C_k\} = 7$, and $b = 180$. Moreover, we employ weighted softmax cross entropy loss against the imbalanced data for training the model by

$$L = -\frac{1}{N_c} \sum_{k}^{N_c} \sum_{c \in C_k} w_c t_c \log(\text{softmax}(y_c)), \tag{17}$$

where $w_c$, $t_c$, and $y_c$ each represent the reciprocal of ratio of the number of samples of the class $c$ to total batch size, ground truth label for the class $c$ (0 or 1), and predicted logit for the class $c$. The weighted softmax cross entropy loss puts more loss on minority classes and results in the avoidance of over-fitting.

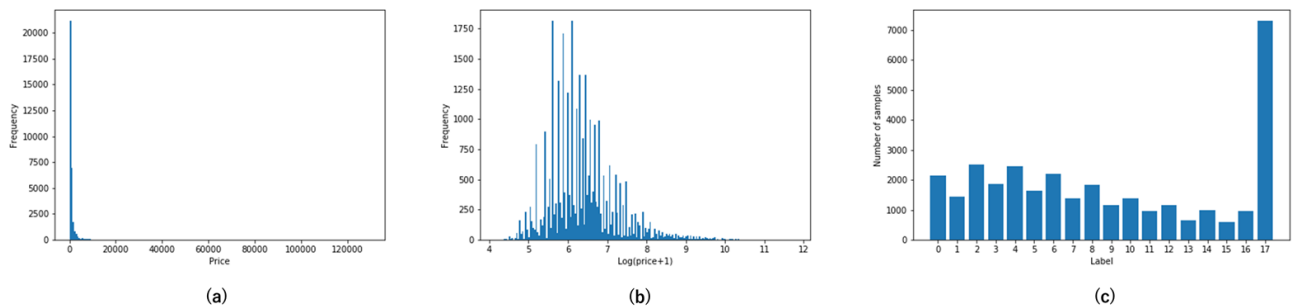

**Figure 4: (a) Long-tail distribution of prices. Prices are written in dollars. (b) Distribution of log-transformed prices. (c) The number of samples for each class. As for the definition of the classes, refer to Table 2. The last class (label:17) covers wide ranges of price zone and contains large number of samples. It is obviously imbalanced data.**

## 5 EXPERIMENTS

### 5.1 DATASET

We have received a large dataset of purchase and selling information of secondhand no brand rings from the collaborative fashion retailer. Our dataset totally includes 32,542 samples with their images and attributes. We split the whole dataset into *train*, *validation*, *test* sets and each includes 19,176, 5,350, and 8,016 samples respectively. The *validation* split is used for hyper-parameter tuning and we employ the model reaching the peak of accuracy on the *validation* split for *test* evaluation. The ring images are size of 512×512. These images are originally taken for the purpose of insertion into the retailer's e-commerce site, and thus the photographic conditions such as the camera viewpoint, illumination, and background are mostly the same among all samples. The details of the attributes are shown in Table 1. Both continuous and categorical variables are included. In addition to these characteristic attributes, we add 'month of pur chase' to take the seasonality into account in our models. We have a total of 13 types of attributes and 123 variables, and we therefore set parameter $T$ introduced in Section 4.2 to 13. As described in Section 3, we use the selling prices for the labels. Figures 4 (a) and (b) show the histogram of the price, which has an extremely long tail distribution. In order to discretize the continuous prices into categorical classes, we defined 18 price zones (labels) along the price line, which is a list of reference selling prices determined by the collaborative retailer. Table 2 shows the 18 price zones and label IDs, where the prices are written in dollars and converted from Japanese yen based on the exchange rate as of May 2019. We also plot the number of samples for each label in Figure 4 (c).

### 5.2 IMPLEMENTATION DETAILS

We use the Momentum Stochastic Gradient Descent optimizer with a learning rate $\gamma = 0.01$. We train models on the *train* split, validate on the *validation* split, and evaluate the performance on the *test* split. The batch size is set to 16 for all models. While training a model, input images are randomly cropped to size of 448×448. As for the attributes, we filled missing values in the continuous variables with their mean value and then we apply a log-transformation and standardization.

### 5.3 EVALUATION METRIC

We evaluate the performances of models by top-$k$ accuracies, where we sort predicted labels by their confidence values in descending order and calculate accuracy according to whether any of $k$ highest labels coincide to the correct label or not. We set $k = 3, 5, 7$ to observe how much the models concentrate their prediction to the correct price.

### 5.4 METHODS

To explore the essential components in multimodal models for a resale price prediction task, we implemented various models in our experiment, which can be roughly divided into the baseline models, state-of-the-art models, and our proposed models. To investigate how each modality affects the price, we built the 'visual-alone' and 'attributes-alone' models as follows: (i) The visual-alone model extracts a visual feature vector of 2,048 dimensions from the 'pool5' layer in ResNet-50, which is fed to the following MCE layer. (ii) The attributes-alone model adds a fully connected layer after the second embedding layer to extract a feature vector of 2,048 dimensions, which is fed to the MCE layer. In addition to these single modality models, we built simple multimodal models to purely confirm the effectiveness of our multimodal approach for a resale price prediction task. We conducted the following operations on the feature vectors of the visual appearance and attributes obtained through the same procedure as the single modality models. (iii) The simple concatenation model of two vectors results in a feature vector of 4,096 dimensions. (iv) In the element-wise addition model, the two feature vectors are added in an element-wise manner. (v) Finally, the element-wise product is the product of the two feature vectors. These models are similarly followed by the MCE layer.

**Table 1: List of attributes of a secondhand no brand ring. Both continuous and categorical variables are included. We have totally 13 attributes and 123 variables.**

| ID | Attribute | Description | Variable type | Examples | # of categories |
|---|---|---|---|---|---|
| 1 | Size | The circumference of the ring. | Continuous | 5 | - |
| 2 | Weight | The total weight of the ring. | Continuous | 0.15 ounce | - |
| 3 | Size of center stone | The size of the stone at the center. | Continuous | 0.5 | - |
| 4 | Size of side stone | The size of a stone to the side of the center stone. | Continuous | 0.5 carat, 0.1 inch | - |
| 5 | Grading report | The existence of a grading report showing an analysis of the 4Cs (carat, color, clarity, cut). This is for diamonds only. | Categorical | Present / absent | 2 |
| 6 | Identification report | The existence of a gem identification report, which shows an analysis of the type of gem and its authenticity. This is for all types of gems. | Categorical | Present / absent | 2 |
| 7 | Center stone | The type of center stone. | Categorical | Diamond, ruby, sapphire, pearl, emerald, topaz, opal, etc. | 37 |
| 8 | Side stone | The type of side stone. | Categorical | Diamond, ruby, sapphire, pearl, emerald, topaz, opal, etc. | 37 |
| 9 | Metal | The color and content of the metal. | Categorical | 10K, 14K, 18K, 24K, yellow gold, platinum, silver, etc. | 11 |
| 10 | Degree of use | The quantized degree of use of the ring. | Categorical | Brand-new, unused, used, etc. | 4 |
| 11 | Notices of repair | Noted resizes and adjustments. | Categorical | Non-resizable, non-adjustable, etc. | 6 |
| 12 | Notices of defects | Noted scratches, flaws, or inclusions. | Categorical | Scratches, flaws, inclusions, etc. | 6 |
| 13 | Month of purchase | The month of purchase. | Categorical | January, February, March, etc. | 12 |

**Table 2: Class label IDs and their pricelines. Prices are shown in dollar.**

| label | Price line | Price zone | label | Price line | Price zone |
|---|---|---|---|---|---|
| 0 | 157 | -180 | 9 | 562 | 540-585 |
| 1 | 202 | 180-225 | 10 | 607 | 585-630 |
| 2 | 247 | 225-270 | 11 | 652 | 630-675 |
| 3 | 292 | 270-315 | 12 | 697 | 675-720 |
| 4 | 337 | 315-360 | 13 | 742 | 720-765 |
| 5 | 382 | 360-405 | 14 | 787 | 765-810 |
| 6 | 427 | 405-450 | 15 | 832 | 810-855 |
| 7 | 472 | 450-495 | 16 | 877 | 855-900 |
| 8 | 517 | 495-540 | 17 | 904 | 900- |

For state-of-the-art models, we implemented several bilinear approaches. (vi) The MFB and (vii) MFH models are Matrix Factorization based bilinear models, as described in Section 4.3.2. We set the parameters $k$ and $o$ to 5 and 200, respectively. For the MFH model, the number of MFB blocks $p$ is set to 2. These parameters are the same as in previous studies [15, 16]. In terms of the tensor decomposition, we implemented (viii) MUTAN and (ix) BLOCK models. For the MUTAN model, we set the parameters $t_q = t_v = t_o = 360$, and rank $R = 10$. These are the same as used in the experiment, the details of which can be found in [17]. In the BLOCK model, we chose $L = M = N = 32$ and $R = 100$, which differ from the experiment settings, which were $L = M = N = 80$ and $R = 20$ for the VQA task [18]; however, we found that the BLOCK model provides better results when making the core tensor smaller and increasing the rank. For the proposed models, we implemented the iterative co-attention networks (hereinafter, the ICAN model). To observe the reactions when changing the number of cells $L$, we built the ICAN models for $L = 1$ to 5. The comparison results of the top-$k$ accuracy are shown in Table 3 and the peak is almost reached when $L = 3$. The results of the ICAN models when (x) $L = 1$ and (xi) $L = 3$ are listed in Table 4. The parameter settings in the MFB and BLOCK modules of the co-attention cell are the same as with the single models.

## 5.5 RESULTS

As shown in Table 4, the attributes-alone model shows better results than the visual-alone model. This result indicates that explicit product information gives steady predictions. The product image also implicitly includes basic information such as the size or type of a stone, but this is ambiguous and results in poor accuracy. By contrast, multimodal models, namely, concatenation, element-wise addition, and element-wise product, are clearly more accurate than single-modality models. We infer that while the attributes provide explicit product information, the images provide visual information that does not appear in the attributes, and therefore these modalities are complementary to each other. Thus, the multimodal approaches are effective for a resale price prediction. The results of the state-of-the-art models, MFB, MFH, MUTAN, and BLOCK, indicate that they all excel over the simple multimodal models. This shows that bilinear approaches are certainly effective in a multimodal fusion. They achieve almost the same results although the BLOCK model slightly excels over the others, and thus we adopted it for the final fusion of our co-attention cell. Our ICAN models remarkably outperform these state-of-the-art models with top-3, -4, -5 level accuracies. In particular, for $L = 3$ ((xi) in Table 4), +6.8– 9.7 and +5.9 point improvements are achieved in top-3 level accuracy over the simple multimodal fusion models ((iii)-(v) in Table 4) and the best bilinear approach, namely, the BLOCK model ((ix) in Table 4), respectively. For $L = 1$ ((x) in Table 4), where the ICAN model corresponds to the BLOCK model in which our co-attention mechanism is applied only once, the accuracy is significantly improved compared to a simple BLOCK model. We also observed that increments in the iterative co-attention process enable a boost in performance ((xi) in Table 4). This result indicates that the attention process should be repeated several times to correctly specify the important parts of the input data and should be alternately interacted between different modalities. It is thought that our iterative co-attention architecture follows the human attention system, namely, when humans act in the real world, the top-down and bottom-up attention systems do not operate individually, but are interactively applied. Figure 5 shows the confusion matrix of the ICAN model when $L = 3$ on *test* split, which indicates that the predictions are concentrated along the diagonal of the matrix; in other words, the predicted class distribution overlaps the correct label. From the evaluation results of multiple models, the collaborative retailer has regarded the ICAN model as achieving a practical level performance.

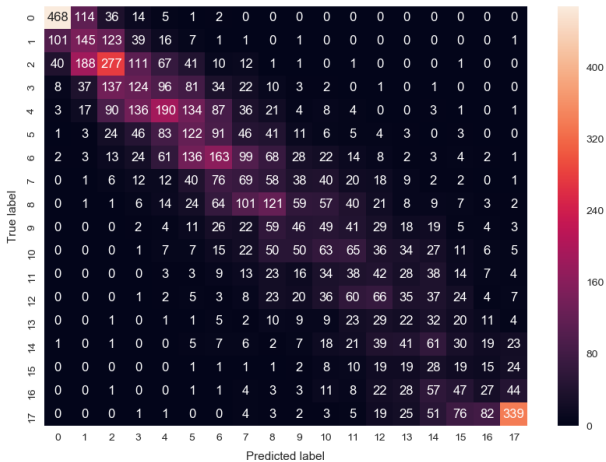

**Figure 5: Confusion matrix of iterative co-attention networks when $L = 3$ on *test* split. Predicted distribution overlaps the correct label and predictions concentrate on the diagonal of matrix.**

We further visualized each attention map for visual and attribute representations, as shown in Figure 6. We picked up several samples using (a) the original product image, (b) a visual attention heatmap of where the model focuses on, (c) an image overlaid with the attention heatmap, and (d) an attribute attention heatmap along with the names of the attributes. From the visual attention map, we can see that the ICAN model reveals important regions of the input image, such as over the center stone, side stone, or other characteristic parts. Similarly, important attributes are highlighted including the weight, size of the center stone, and size of the side stone. These factors are intuitively understandable, but we found that the center stone is not as highlighted as we expected, and this information might be retrieved from the input image.

In addition, we interestingly observed that an identification report is occasionally highlighted. Since its existence does not appear in images, this indicate that the model might extract modality-dependent factors to complement each other. For several products whose characteristics make it extremely difficult to estimate their prices, the attention heatmaps cover the entire appearance of a product and its attributes to seek clues to estimate its price. Moreover, our model provides new insights into further analysis. We found that the month of purchase is often highlighted, which indicates there may be a seasonality in the resale prices. Perhaps, anniversary promotions or trading among resale retailers affect their pricing trend. Thus, a co-attention mechanism can discover new information and provide areas of further investigation.

**Table 3: Top-$k$ accuracy of iterative co-attention networks when changing the number of cells. It is peaked around $L = 3$.**

| Models | Top-$k$ accuracy [%] | | |
|---|---|---|---|
| | $k = 3$ | $k = 5$ | $k = 7$ |
| ICAN with $L = 1$ | 57.8 | 77.1 | 88.6 |
| ICAN with $L = 2$ | 58.7 | 77.3 | 88.3 |
| ICAN with $L = 3$ | **59.0** | **77.6** | 88.8 |
| ICAN with $L = 4$ | 58.6 | 77.4 | **88.9** |
| ICAN with $L = 5$ | 57.6 | 77.1 | 88.7 |

**Table 4: Evaluation results of top-$k$ accuracies on various models. We implement baseline models and state-of-the-art models in multimodal fusion as well as our proposed iterative co-attention networks. The ICAN models outperform both baseline and state-of-the-art models.**

| Models | Top-$k$ accuracy [%] | | |
|---|---|---|---|
| | $k = 3$ | $k = 5$ | $k = 7$ |
| (i) visual-alone | 41.6 | 59.9 | 73.6 |
| (ii) attributes-alone | 46.6 | 66.0 | 78.9 |
| (iii) concatenation | 50.4 | 69.9 | 82.8 |
| (iv) element-wise addition | 49.3 | 69.0 | 82.6 |
| (v) element-wise product | 52.2 | 71.7 | 84.3 |
| (vi) MFB [15] | 52.3 | 71.7 | 85.1 |
| (vii) MFH [16] | 52.9 | 73.5 | 85.5 |
| (viii) MUTAN [17] | 52.8 | 72.8 | 85.7 |
| (ix) BLOCK [18] | 53.1 | 73.9 | 86.4 |
| (x) ICAN with $L = 1$ | 57.8 | 77.1 | 88.6 |
| (xi) ICAN with $L = 3$ | **59.0** | **77.6** | **88.8** |

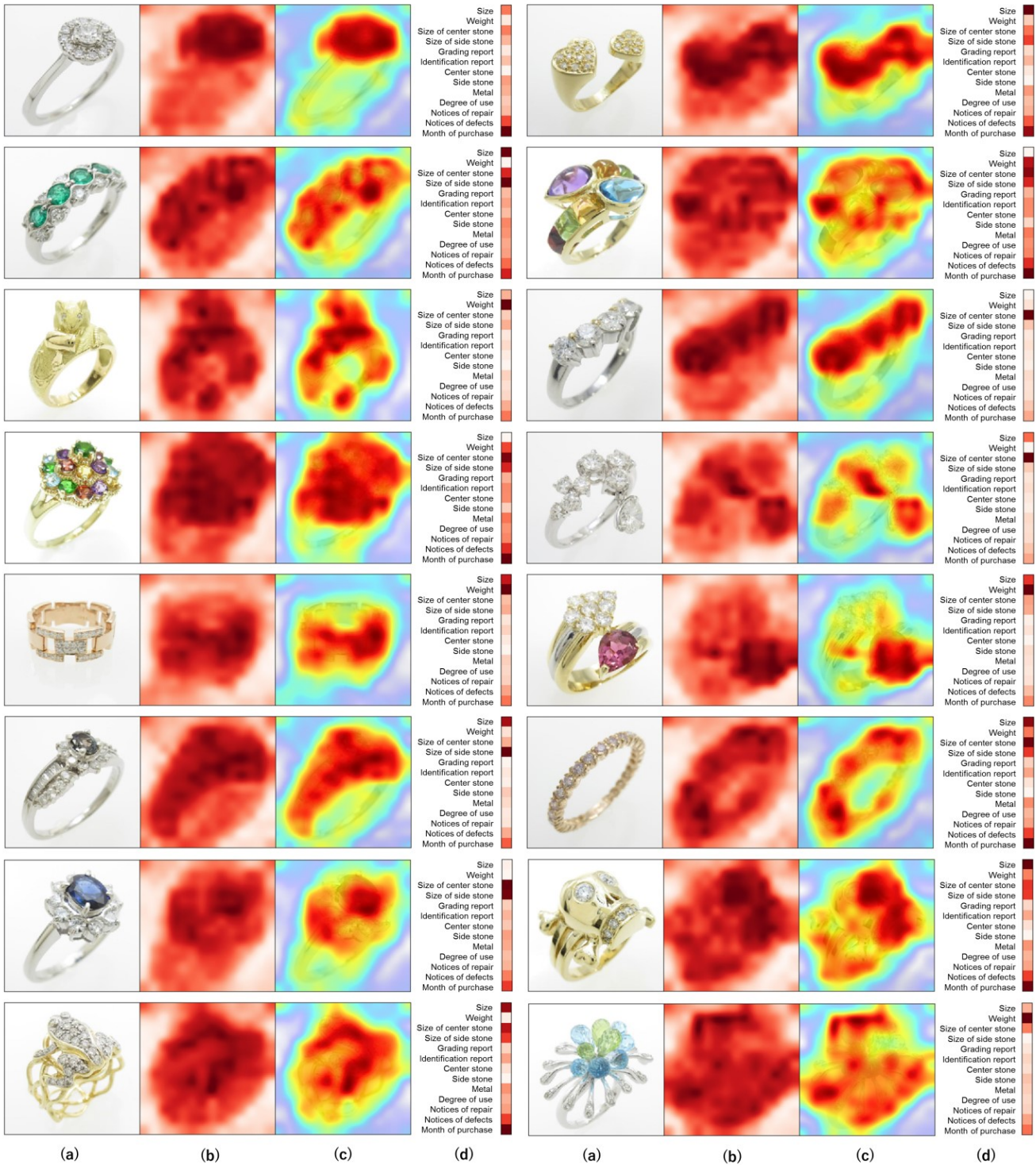

Figure 6: Visualization results of attention maps for an image and attributes data: (a) original product image, (b) visual attention heatmaps of where the model focuses on, (c) overlaid images with attention heatmaps, and (d) attention heatmaps of attributes along with the names. Visually important parts such as the center stone, side stone, and other characteristic parts are highlighted in the overlaid images. Similarly, important attributes are highlighted such as the weight, size of the center stone, and size of the side stone. In the case of unusual design samples (for example, bottom row, left), attention maps cover the whole range of the input data; in other words, the model seeks clues for a resale price prediction from the input data. Such visualization helps users understand the prediction results.

## 5.6 DISCUSSIONS AND FUTURE WORKS

We demonstrated that our iterative co-attention model is effective in the resale price assessment of secondhand jewelry items, and showed that it achieves a high level of performance experimentally and hence the collaborative retailer has regarded the ICAN model as achieving a practical level performance. However, we need to apply additional experiments to confirm its practicality. First, we need to conduct an A/B test, where half of the target products are priced by experts and the other half are priced by our system. The sales and inventory turnover can then be fairly compared. For the model versatility, we need to seek a methodology to optimize multiple hyper-parameters in our model. In addition, we should incorporate the trends of the economic conditions, the market prices of the stones and metals, or the design popularity. We will extend our iterative co-attention model to other fashion items or tasks, as well as other domains, where the appearance design and specifications are both important.

## 6 CONCLUSION

In this paper, we introduced a resale price prediction task of secondhand jewelry items and presented a multimodal model that takes an image and attributes of a product into consideration, and thus suggests a relevant resale price. We employed state-of-the-art methods in VQA, which is a typical multimodal task, and proposed a new iterative co-attention network that models an expert's pricing procedure in which the appearance and specifications of a product are carefully and iteratively observed. In our experiment on the large dataset of secondhand no brand rings, we demonstrated that our model provides an outstanding performance compared to baseline and state-of-the-art fusion models. Furthermore, we clarified the key factors of the input multimodal data by visualizing attention maps for each modality using a co-attention mechanism. Our model architecture is also applicable to other fashion items, tasks, and domains, where the design appearance and specifications are both important.